\documentclass[conference]{IEEEtran}
\IEEEoverridecommandlockouts
\usepackage{cite}
\usepackage{amsmath,amssymb,amsfonts}
\usepackage{algorithmic}
\usepackage{graphicx}
\usepackage{textcomp}
\usepackage{dirtytalk}
\usepackage{xcolor}
\usepackage{multirow}
\usepackage{algorithm}
\usepackage{hyperref}

\def\BibTeX{{\rm B\kern-.05em{\sc i\kern-.025em b}\kern-.08em
    T\kern-.1667em\lower.7ex\hbox{E}\kern-.125emX}}
\begin{document}

\title{Artificial Interrogation for Attributing Language Models\\
{\footnotesize \textsuperscript{*}National University of Computer and Emerging Sciences (FAST-NUCES), Karachi, Pakistan}
\thanks{Identify applicable funding agency here. If none, delete this.}
}

\author{\IEEEauthorblockN{1\textsuperscript{st} Farhan Dhanani}
\IEEEauthorblockA{
OCRID: 0000-0002-7490-7655\\
\textit{Dept. of Computer Science}}
\textit{Ph.D. Student}\\
\and

\IEEEauthorblockN{2\textsuperscript{nd} Muhammad Rafi}
\IEEEauthorblockA{
OCRID: 0000-0002-3673-5979\\
\textit{Dept. of Computer Science}}
\textit{Associate Professor}\\

}

\maketitle

\begin{abstract}
This paper presents solutions to the Machine Learning Model Attribution challenge (MLMAC) collectively organized by MITRE, Microsoft, Schmidt-Futures, Robust-Intelligence, Lincoln-Network, and Huggingface community. The challenge provides twelve open-sourced base versions of popular language models developed by well-known organizations and twelve fine-tuned language models for text generation. The names and architecture details of fine-tuned models were kept hidden, and participants can access these models only through the rest APIs developed by the organizers. Given these constraints, the goal of the contest is to identify which fine-tuned models originated from which base model. To solve this challenge, we have assumed that fine-tuned models and their corresponding base versions must share a similar vocabulary set with a matching syntactical writing style that resonates in their generated outputs. Our strategy is to develop a set of queries to interrogate base and fine-tuned models. And then perform one-to-many pairing between them based on similarities in their generated responses, where more than one fine-tuned model can pair with a base model but not vice-versa. We have employed four distinct approaches for measuring the resemblance between the responses generated from the models of both sets. The first approach uses evaluation metrics of the machine translation, and the second uses a vector space model. The third approach uses state-of-the-art multi-class text classification, Transformer models. Lastly, the fourth approach uses a set of Transformer based binary text classifiers, one for each provided base model, to perform multi-class text classification in a one-vs-all fashion. This paper reports implementation details, comparison, and experimental studies, of these approaches along with the final obtained results.
\end{abstract}


\section{Introduction}
The last decade has revealed increased growth in the development of large language models to solve diverse problems of the practical world that were previously un-comprehended for machines. The advancements in language models and the public release of GPT\cite{GPT} models have blurred the lines between reality and artificial simulation and also raised concerns about its responsible usage. Big tech companies who own expensive hardware train these models and ensure their reliability, real-time capability, and effective usage. The big companies carefully craft these huge language models to restrict the model from giving any racist responses or hurting the sentiments of people in any circumstances before allowing large public access to these models. When these models are openly accessible to the public, anyone can cheaply fine-tune these models on attractive domain-specific topics such as politics to earn benefits or even misuse by fine-tuning them on malicious text. Tracking such activities requires sophisticated methods to be developed and deployed to catch the wrongdoers before any disastrous outcome or financial loss. The Machine Learning Model Attribution challenge\cite{mlmac} (MLMAC) addresses this subject by calling lead researchers from the industry to pair a given unnamed fine-tuned model with the most probable model from the provided list of base models, given that the models can only access via provided REST APIs. The contest introduces two sets of models, each containing twelve models. The first set consists of fine-tuned models, and the second set includes base models. The task for participants is to design an algorithmic technique to pair each fine-tuned model with its corresponding base model. It is important to note that there can be a fine-tuned model from the first set that does not originates from any of the provided base models in the second set. Or it can also be a case where a base model is used to produce more than one fine-tuned model. The contest openly announces the names of the base models, and participants are free to check their implementation, as all of them are publically accessible. However, the participants do not have access to the names and architectural structures of the fine-tuned models. Participants can only refer them through numbers and access them via API provided by the organizers. For example, the fine-tuned model number 1, fine-tuned model number 2, and so on. To solve this task, we have assumed that the fine-tuned model and its corresponding base model should share a common set of vocabulary. Plus, both models should also demonstrate similarities in the use of words and phrases in their generated outputs for any given query. Based on this assumption, we have sampled ninety queries from ten publicly available datasets and developed four distinct approaches to solve this task. We have recorded the responses of all models for the collected queries and then prepared four different solutions to measure the similarities in the outputs generated by the fine-tuned and base models. The first approach uses BLEU\cite{papineni-etal-2002-bleu} and TER\cite{TER} scores to estimate the resemblance in the responses of base and fine-tuned models. The second approach uses a vector space model\cite{vsm} where a similarity score gets calculated between the response produced by the fine-tuned model and the outputs of the base models for a given query. The process is repeated hundred times for each distinct query, and then the base model with the most similar responses gets selected. The third approach uses state-of-the-art multi-class text classification, Transformer\cite{trans} models. They take the textual sequence produced from a fine-tuned model as their input and emit the name of the base model as a category label in the output. It's essential to note that all three approaches discussed here can not identify the case when the given fine-tuned model doesn't belong to any of the provided base models in a self-supervised fashion. Our last proposed method addresses this issue by training separate binary sequence classifiers for each base model in the one-vs-all multi-class classification style for producing a binary decision. The binary decision entails whether its associated base model can generate the given input sequence or not, based on the used vocabulary terms and their ordering. We pass the output of a fine-tuned model for a given query as an input to all these binary sequence classification models, and then select the decision of the binary sequence classifier that emits a maximum score entailing its associated base model can generate the given input sequence. If all the models infer that their corresponding base model can not produce the output of the fine-tuned model, then we can conclude that the given fine-tuned model can not be paired with any of the provided base models.

\section{Materials and Methods}
This section presents an overview of our techniques and outlines the motivation for selecting these approaches to solve the Machine Learning Model Attribution challenge\cite{mlmac} (MLMAC).

\subsection{Assembling Queries for Interrogating Models}
\begin{figure}[htbp]
\centerline{\includegraphics[width=0.38\textwidth]{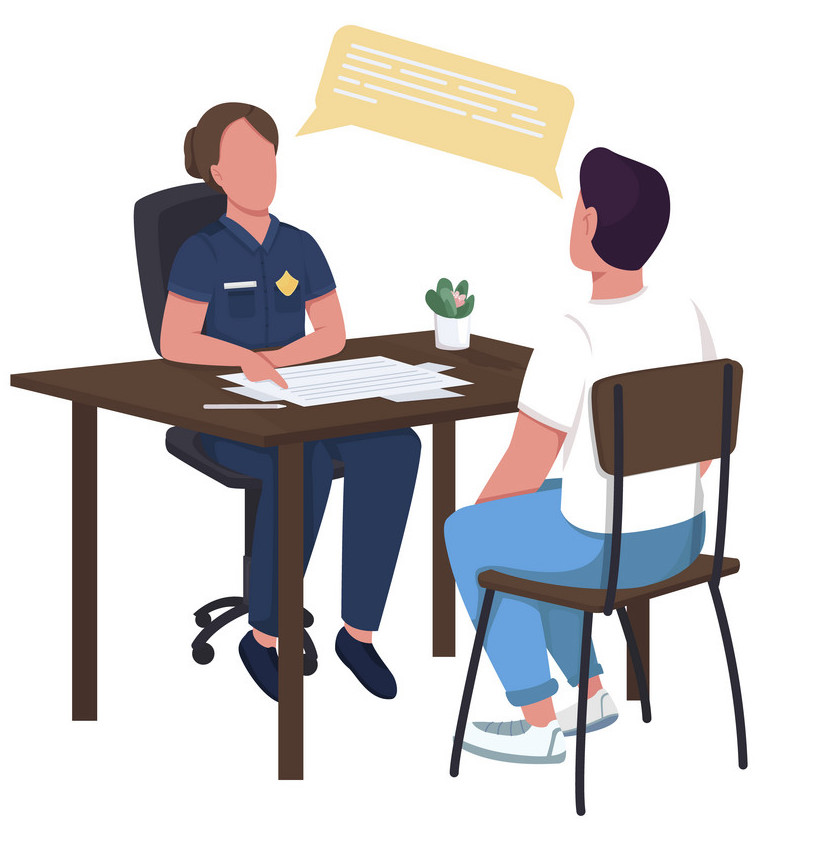}}
\caption{The image visually picturizes the scenario in which an investigator is interrogating a suspect to uncover the truth.}
\label{fig1}
\end{figure}

Imagine you are working as a senior investigator for a reputed agency and interrogating a group of prime suspects to identify the culprit of a serious crime, as illustrated in figure \ref{fig1}. You may need to consider multiple possibilities before accusing any suspects and finalizing your conclusion. It can be a case where none of them is a culprit, or maybe one of them is a culprit. Another possibility can be that an individual from the group has organized the whole plan to carry out the mission. And either few of the remaining suspects or all of the suspects have supported the culprit. Your job is to extract true confessions from the culprits involved in the crime and uncover their names to solve the case. In such a critical situation, what tactics do you choose to unveil the truth? Many psychologists believe that coercive or confrontational methods do not yield ideal outcomes, because many times the circumstances do not allow you to demonstrate hostile behavior. Moreover, in some cases, aggressive actions also impair the senses of the suspect to accept the truth, which eventually complicates your work. Think for a while if you can not forcefully retrieve the confessions, then what other options do you have to discover the truth?

Experts think the best tactic to reveal the truth in such foggy situations must rely on framing intelligent queries that offer temporary rewards to the culprits, and provoke them to answer honestly. However, in the long-term the sequence of answered facts tricks the culprits by un-consciously unveiling their true intentions and the hidden truth. The \text{\say{Prisoner's Dilemma}}\cite{Prisoners_dilemma} problem is a good case study demonstrating the discussed tactic. In the MLMAC\cite{mlmac} challenge, the participants faced a similar situation where all the provided models were language models, and we can only access them via APIs. The fine-tuned models will not simply reveal their corresponding base model and vice versa. Therefore, to find the original base models of the provided fine-tuned models, we have articulated a similar tactic and gathered a list of queries to solve the challenge. Our strategy exploits the fundamental nature of fine-tuning to find the most probable base model for a given fine-tuned model from a list of base models after considering their responses. The art of fine-tuning is to tailor the outputs of a pre-trained base model for specific topics to increase its usefulness in a particular domain. For this reason, we believe that by ensuring topic-wise diversity among the collected queries, we can provoke fine-tuned models for generating similar responses compared to their base versions because if a base model gets fine-tuned for a domain \textit{\say{x}}, and we query the fine-tuned model on the same domain, then its response will substantially differ from its corresponding base version. Conversely, the output of the resultant fine-tuned model will be theoretically much more similar to its corresponding base version if we query it outside the domain \textit{\say{x,}} on which it wasn't originally fine-tuned. In this challenge, we are unaware of which fine-tuned model gets customized for which domain. Therefore, we think the more diverse queries we present to the fine-tuned models, the higher the chances that the posed query doesn't belong to the domain on which the model has been fine-tuned. And this will eventually increase the prospects of spotting similarities in the generated responses from the base model and its fine-tuned version. For example, we can easily use a  pre-trained GPT-x model and fine-tune it on the medical data set to facilitate medical students in extracting medical information from a large data set. Given such a scenario, it will be hard to observe the similarity in the generated outputs of the GPT-x model and its fine-tuned version for a given query that belongs to the medical domain. However, there will be higher chances of observing a similarity in the produced outputs from both models if we pose a query from the literature domain. Hence, the diversity among the used queries plays a vital role in attributing fine-tuned models with their base versions via interrogation. It is also essential to mathematically quantify this idea of diversity for measuring the goodness of the collected set of queries, and the following equation \ref{eq1} accomplishes this job.
\begin{equation}
\begin{split}
Diversity =& \quad \frac{\textit{No. of Datasets}}{\textit{No. of Queries }} \times 100\\
\textit{where }\rightarrow & \textit{No. of Datasets = Number of text datasets on}\\
& \textit{different topics from which queries are}\\
& \textit{getting sampled.}\\
\rightarrow & \textit{No. of Queries = Number of sampled queries.}
\end{split}
\label{eq1}
\end{equation}
\begin{table}[h!]
\center
\begin{tabular}{ ||c|p{7cm}||  } 
 \hline
 \textbf{No. } & \multicolumn{1}{|c||}{\textbf{Dataset}} \\  
 \hline\hline

 1 & A Multi-Task Benchmark and Analysis Platform for Natural Language Understanding \cite{wang2019glue}\\
 \hline
 2 & Seeing stars: Exploiting class relationships for sentiment categorization with respect to rating scales \cite{Pang+Lee:05a}\\
 \hline
 3 & SQuAD: 100,000+ Questions for Machine Comprehension of Text \cite{2016arXiv160605250R}\\
 \hline
 4 & Learning Word Vectors for Sentiment Analysis \cite{maas-EtAl:2011:ACL-HLT2011}\\
 \hline
 5 & Learning Question Classifiers \cite{li-roth-2002-learning}\\ 
 \hline
 6 &  No Language Left Behind: Scaling Human-Centered MT\cite{https://doi.org/10.48550/arxiv.2207.04672}\\ 
 \hline
 7 & A Monolingual Approach to Contextualized Word Embeddings for Mid-Resource Languages \cite{ortiz-suarez-etal-2020-monolingual}\\
 \hline
 8 & RACE: Large-scale Reading Comprehension Dataset From Examinations \cite{lai2017large}\\
 \hline
 9 & BLiMP: A Benchmark of Linguistic Minimal Pairs for English\cite{warstadt2019blimp}\\
 \hline
 10 & PIQA: Reasoning about Physical Commonsense in NL\cite{Bisk2020}\\ 
 \hline
 \hline
\end{tabular}
\caption{The table lists the collection of datasets that we have used to sample the queries for interrogating all the provided language models in the MLMAC\cite{mlmac} challenge.}
\label{table:1}
\end{table}

Here, equation \ref{eq1} defines diversity as the ratio of the number of used text datasets for sampling the queries with the number of total sampled queries. The value of the diversity will be minimum when only one dataset gets used for sampling all the queries. Conversely, the value of the diversity will be maximum if each query gets sampled from a new data set. In our experiments to solve the MLMAC\cite{mlmac} challenge, we randomly selected nine queries from ten different data sets, as listed in table \ref{table:1}. Each data set belongs to a unique topic, and we have extracted a total of ninety queries from them to interrogate all the provided language models in the contest.

\subsection{APPROACH-1: Machine Translation Evaluation metrics}
BLEU \cite{papineni-etal-2002-bleu} and TER \cite{TER} scores are non-differential automatic machine translation evaluation metrics. They are used to rate the quality of produced translations from artificial machines (or deep learning models) by comparing them with the reference translations annotated by humans. These evaluation metrics mathematically quantify the similarity between the produced translations by the machines with the human-annotated translations. We have utilized this characteristic of BLEU \cite{papineni-etal-2002-bleu} \& TER  \cite{TER} scores for pairing fine-tuned language models with their corresponding base models by finding the similarity in their generated responses for a given set of queries, as explained in Algorithm \ref{alg1} below.

\begin{algorithm}
\caption{Attributing Fine-tuned Language Models with Base Models VIA BLEU \& TER Scores}
\label{alg1}
\begin{algorithmic} 
\REQUIRE $list\_of\_fine\_tuned\_models,$\newline \hspace*{3em}$list\_of\_base\_models, list\_of\_queries$
\ENSURE $len(list\_of\_queries)>0$ \&\\
\hspace*{3em}$len(list\_of\_base\_models)>0$ \&\\
\hspace*{3em}$len(list\_of\_fine\_tuned\_models)>0$
\STATE $t\_pairs \leftarrow \{\}$
\STATE $b\_pairs \leftarrow \{\}$
\FOR{$fm \;$ in $\;list\_of\_fine\_tuned\_models$}
\STATE $avg\_bleu\_sc \leftarrow []$
\STATE $avg\_ter\_sc \leftarrow []$
\FOR{$bm \;$ in $\;list\_of\_base\_models$}
\STATE $bleu\_sc \leftarrow []$
\STATE $ter\_sc \leftarrow []$
\FOR{$query \;$ in $\;list\_of\_queries$}
\STATE $b\_resp \leftarrow bm(query)$
\STATE $ft\_resp \leftarrow fm(query)$
\STATE $bleu\_score \leftarrow calc\_BLEU(b\_resp, ft\_resp)$
\STATE $ter\_score \leftarrow calc\_TER(b\_resp, ft\_resp)$
\STATE $bleu\_sc.append(bleu\_score)$
\STATE $ter\_sc.append(ter\_score)$
\ENDFOR
\STATE $avg\_bleu\_sc.append(sum(bleu\_sc)/len(bleu\_sc))$
\STATE $avg\_ter\_sc.append(sum(ter\_sc)/len(ter\_sc))$
\ENDFOR
\STATE $t\_pairs[fm] \leftarrow list\_of\_base\_models.index($\\
\hspace*{6.4em}$avg\_ter\_sc.index (min(avg\_ter\_sc)))$
\STATE $b\_pairs[fm] \leftarrow list\_of\_base\_models.index($\\
\hspace*{6.4em}$avg\_bleu\_sc.index (max(avg\_bleu\_sc)))$
\ENDFOR
\end{algorithmic}
\end{algorithm}

The algorithm \ref{alg1} outlines all the details to pair the given fine-tuned models with the provided base models via simulated interrogation. One important thing to note is that the higher the BLEU\cite{papineni-etal-2002-bleu} score exists between two given responses, the more they are alike. However, the lower the TER\cite{TER} score exists between two given responses, the more they are similar.

\subsection{APPROACH-2: Vector Space Models}
The applications of  vector space models\cite{vsm} are popular in information retrieval systems, where they are used to measure the similarity of a given query with the collection of documents and rank the documents based on their relevance. However, the basic idea of vector space models is to articulate a general framework that can be useful in any problem to calculate a similarity score between two vectors from a geometrical viewpoint. We have applied the VSM\cite{vsm} model to find the most suitable base version from the list of base models for a given fine-tuned model by finding the similarities in their generated responses on a set of input queries, as shown in algorithm \ref{alg2}.
\begin{algorithm}
\caption{Attributing Fine-tuned Language Models with Base Models VIA Vector Space Models}
\label{alg2}
\begin{algorithmic} 
\REQUIRE $list\_of\_fine\_tuned\_models,$\newline \hspace*{3em}$list\_of\_base\_models, list\_of\_queries$
\ENSURE $len(list\_of\_queries)>0$ \&\\
\hspace*{3em}$len(list\_of\_base\_models)>0$ \&\\
\hspace*{3em}$len(list\_of\_fine\_tuned\_models)>0$
\STATE $docs \leftarrow \{\}$
\FOR{$bm \;$ in $\;list\_of\_base\_models$}
\STATE $all\_resp \leftarrow$ \say{}
\FOR{$query \;$ in $\;list\_of\_queries$}
\STATE $b\_resp \leftarrow bm(query)$
\STATE $all\_resp\;=\; all\_resp\; +\; b\_resp$
\ENDFOR
\STATE $docs[bm] \leftarrow all\_resp$
\ENDFOR
\STATE $pairs \leftarrow \{\}$
\STATE $sim\_metric \leftarrow cosine\_function$
\STATE $vsm\_model \leftarrow VSM\_MODEL(docs, sim\_metric)$
\STATE $output\_vec\_dim \leftarrow len(list\_of\_base\_models)$
\FOR{$fm \;$ in $\;list\_of\_fine\_tuned\_models$}
\STATE $aggregator \leftarrow vec\_of\_zeros(output\_vec\_dim)$
\FOR{$query \;$ in $\;list\_of\_queries$}
\STATE $f\_resp \leftarrow fm(query)$
\STATE $docs\_sim \leftarrow vsm\_model.compute\_sim(f\_resp)$
\STATE $aggregator = aggregator + docs\_sim$
\ENDFOR
\STATE $aggregated\_avg \leftarrow aggregator / len(list\_of\_queries)$
\STATE $pairs[fm] \leftarrow list\_of\_base\_models.index($\\
\hspace*{6.4em}$aggregated\_avg.index ($\\
\hspace*{7.4em}$max(aggregated\_avg)))$
\ENDFOR
\end{algorithmic}
\end{algorithm}

The algorithm \ref{alg2} sketches the details of our strategy to use the VSM\cite{vsm} framework in solving the MLMAC\cite{mlmac} challenge. First, we have created a response document for each base model by concatenating all of its responses for a given set of input queries. We have treated these response documents as source documents in the VSM\cite{vsm} framework. Next, for each fine-tuned model, we have accumulated the average relevance scores for each document over the responses they have generated for all the given input queries. In the end, we have paired all the fine-tuned models with the base models whose response documents were most similar to their generated responses on average.

\subsection{APPROACH-3: Multiclass Text Classification}
We have mapped the challenge of attributing fine-tuned models with their base version to the problem of multi-class text classification. As all the models in the challenge are language models, we can treat each base model as an output class and all the textual content generated by it as a set of records in that class. In such a way, we can develop a tabular training dataset with three columns, where each row contains three values: the name of the base model, the query provided in the input of the base model, and the response generated by the model for the given query. Then we can train popular deep-learning multi-class text classifiers on the developed dataset to process a piece of text in its input and predict the name of the most probable base model that may have generated it. Next, we can feed the same queries to the fine-tuned models and propagate their generated responses to the trained multi-class text classifier model to infer the most probable base model that may have produced the incoming response. 
\begin{figure}[htbp]
\centerline{\includegraphics[width=0.5\textwidth]{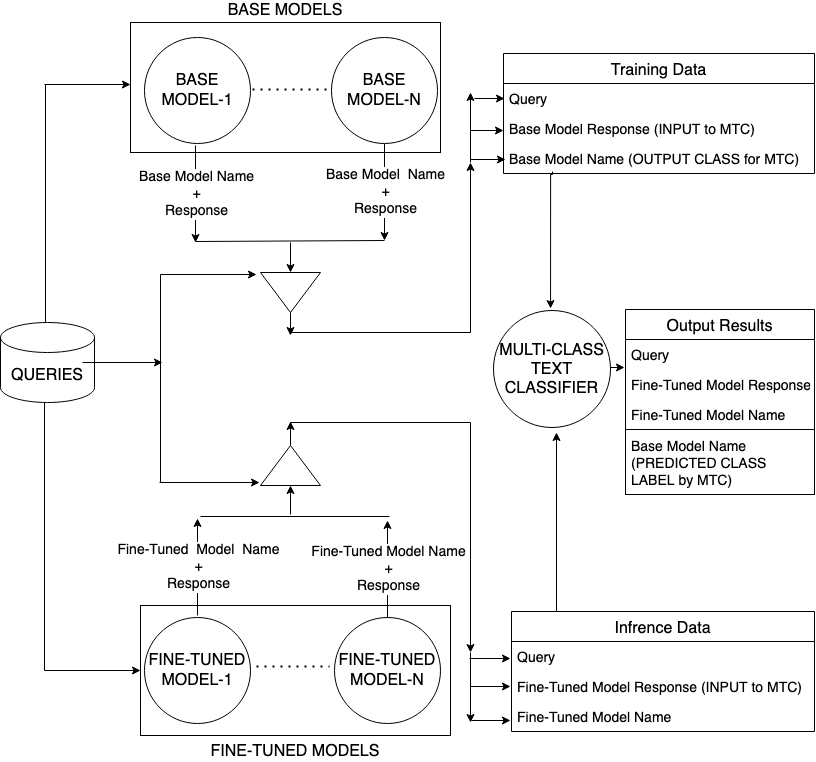}}
\caption{The image presents the architectural structure for mapping the attribution of linguistic fine-tuned models to their corresponding base models on the problem of multi-class text classification.}
\label{fig2}
\end{figure}

Algorithm \ref{alg3}, presented next, outlines our approach for solving the MLMAC\cite{mlmac} challenge with multi-class text classification. The latest transformer-based\cite{trans} models are well-known for delivering state-of-the-art results in solving multi-class text classification problems. Because of their stunning performance and increasing popularity, we have used them as a multi-class text classification model in our implementation.
\begin{algorithm}
\caption{Attributing Fine-tuned Language Models with Base Models VIA Multi-class Text Classification Models}
\label{alg3}
\begin{algorithmic} 
\REQUIRE $list\_of\_fine\_tuned\_models,$\newline \hspace*{3em}$list\_of\_base\_models, list\_of\_queries$
\ENSURE $len(list\_of\_queries)>0$ \&\\
\hspace*{3em}$len(list\_of\_base\_models)>0$ \&\\
\hspace*{3em}$len(list\_of\_fine\_tuned\_models)>0$
\STATE $docs \leftarrow \{\}$
\FOR{$bm \;$ in $\;list\_of\_base\_models$}
\STATE $all\_resp \leftarrow []$ 
\FOR{$query \;$ in $\;list\_of\_queries$}
\STATE $b\_resp \leftarrow bm(query)$
\STATE $all\_resp.append(b\_resp)$
\ENDFOR
\STATE $docs[bm] \leftarrow all\_resp$
\ENDFOR

\STATE $pairs \leftarrow \{\}$
\STATE $number\_of\_epocs = 3$
\STATE $cost\_func \leftarrow categorical\_crossentropy$
\STATE $output\_vec\_dim \leftarrow len(list\_of\_base\_models)$
\STATE $mtc\_model \leftarrow MULTI\_CLASS\_TEXT\_$\\
\hspace*{4em}$CLASSIFICATION\_MODEL(cost\_func)$
\STATE $mtc\_model.fit(docs, number\_of\_epocs)$

\FOR{$fm \;$ in $\;list\_of\_fine\_tuned\_models$}
\STATE $aggregator \leftarrow vec\_of\_zeros(output\_vec\_dim)$
\FOR{$query \;$ in $\;list\_of\_queries$}
\STATE $f\_resp \leftarrow fm(query)$
\STATE $logits \leftarrow mtc\_model.predict(f\_resp)$
\STATE $aggregator = aggregator + logits$
\ENDFOR
\STATE $aggregated\_avg \leftarrow aggregator / len(list\_of\_queries)$
\STATE $pairs[fm] \leftarrow list\_of\_base\_models.index($\\
\hspace*{6.4em}$aggregated\_avg.index ($\\
\hspace*{7.4em}$max(aggregated\_avg)))$
\ENDFOR
\end{algorithmic}
\end{algorithm}
\subsection{APPROACH-4: One-VS-All Multiclass Text Classification}
There are two main problems in using single multi-class text classification model discussed in the previous section for attributing fine-tuned language models with their base version. The first and foremost problem is the difficulty in scaling the multi-class text classifier model. Each time a renowned organization releases a new base model, we need to retrain the multi-class text classifier from scratch to incorporate the semantical knowledge and vocabulary structure for the newly released base model in our solution. The second problem with the single multi-class text classifier model is the inability to identify the inputs that do not belong to any of the given classes in a self-supervised fashion. Thus, to address these problems, we can split the implementation of multi-class text classification into binary text classifications per each individual class using the one-vs-all \cite{one_vs_all} strategy. In our case, we can train a binary text classifier for each of the provided base models. The sole purpose of each binary text classifier is to yield probabilistic scores to quantify its confidence that the given input text and its vocabulary are generated from its associated base model, as explained in algorithm \ref{alg4}.
\begin{algorithm}
\caption{Attributing Fine-tuned Language Models with Base Models VIA One-VS-All strategy}
\label{alg4}
\begin{algorithmic} 
\REQUIRE $list\_of\_fine\_tuned\_models,$\newline \hspace*{3em}$list\_of\_base\_models, list\_of\_queries$
\ENSURE $len(list\_of\_queries)>0$ \&\\
\hspace*{3em}$len(list\_of\_base\_models)>0$ \&\\
\hspace*{3em}$len(list\_of\_fine\_tuned\_models)>0$
\STATE $docs \leftarrow \{\}$
\FOR{$bm \;$ in $\;list\_of\_base\_models$}
\STATE $all\_resp \leftarrow []$ 
\FOR{$query \;$ in $\;list\_of\_queries$}
\STATE $b\_resp \leftarrow bm(query)$
\STATE $all\_resp.append(b\_resp)$
\ENDFOR
\STATE $docs[bm] \leftarrow all\_resp$
\ENDFOR
\STATE $btc\_models \leftarrow \{\}$
\STATE $number\_of\_epocs = 3$
\STATE $cost\_func \leftarrow binary\_crossentropy$
\FOR{$bm \;$ in $\;list\_of\_base\_models$}
\STATE $train \leftarrow \{\}$
\STATE $train[0] \leftarrow docs[bm]^\complement \textit{ //-ve eg. (complement docs[bm])}$
\STATE $train[1] \leftarrow docs[bm]\textit{ //+ve eg. (docs[bm])}$
\STATE $btc\_model \leftarrow BINARY\_TEXT\_$\\
\hspace*{3.1em}$CLASSIFICATION\_MODEL(cost\_func)$
\STATE $btc\_model.fit(train, number\_of\_epocs)$
\STATE $btc\_models[bm] \leftarrow btc\_model$
\ENDFOR
\STATE $pairs \leftarrow \{\}$
\FOR{$fm \;$ in $\;list\_of\_fine\_tuned\_models$}
\STATE $pred\_freq \leftarrow []$
\FOR{$query \;$ in $\;list\_of\_queries$}
\STATE $f\_resp \leftarrow fm(query)$
\STATE $positive\_probs \leftarrow []$
\FOR{$bm \;$ in $btc\_models.keys$}
\STATE $btc\_model \leftarrow btc\_models[bm]$
\STATE $logits \leftarrow btc\_model.predict(f\_resp)$
\STATE $prob = logits[0]\ge logits[1]? -1:logits[1]$
\STATE $positive\_probs.append(prob)$
\ENDFOR
\STATE $pred= max(positive\_probs)$

\STATE $pred\_freq.append(pred==-1?-1:$\\\hspace*{8em}$positive\_probs.index(pred))$

\ENDFOR
\STATE $bmi= mode(pred\_freq)$
\STATE $pairs[fm] \leftarrow bmi==-1? \;\{\}:$\\ \hspace*{8em}$list\_of\_base\_models.index(bmi)$
\ENDFOR
\end{algorithmic}
\end{algorithm}

\begin{figure}[htbp]
\centerline{\includegraphics[width=0.5\textwidth]{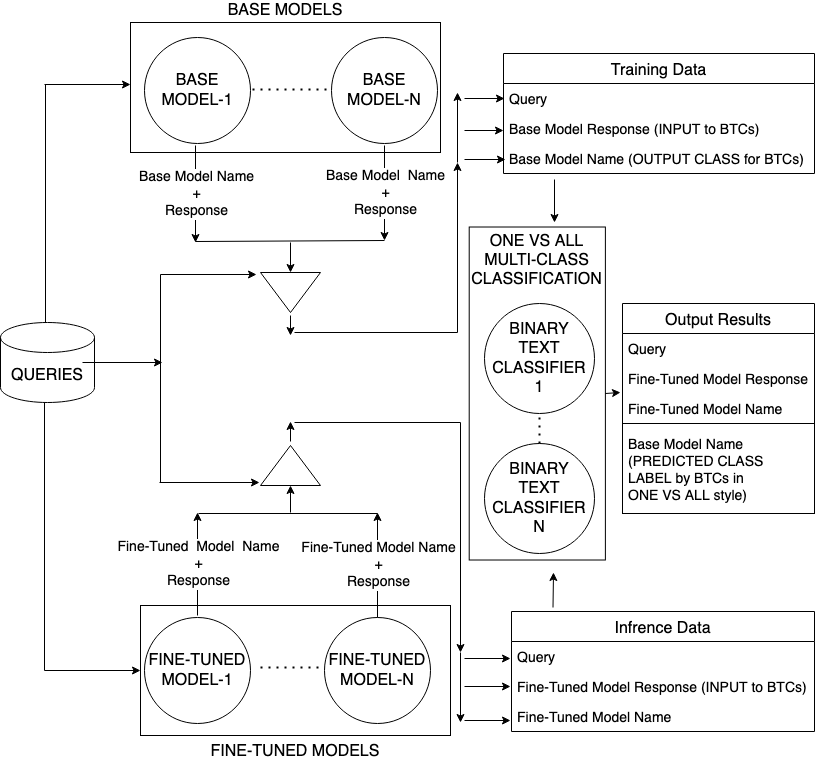}}
\caption{The image illustrates the implementation of a one-vs-all multi-class text classification approach to solve the attribution of fine-tuned language models with base models.}
\label{fig3}
\end{figure}

Figure \ref{fig3} presents an overview of this approach. The overall architecture is almost similar to figure \ref{fig2}, shown in the previous section. The only difference here is that instead of using one multi-class text classifier model, we are using multiple binary classifiers and a collective voting technique for implementing the one-vs-all \cite{one_vs_all} strategy, as described in the algorithm \ref{alg4}. It is important to note that if all binary classifiers predict that the produced response from a fine-tuned model for a given query does not belong to their associated base model, and this inference is repeated for the majority number of collected queries. Only then we conclude that provided fine-tuned model does not belong to any of the given base models.

\section{Results And Experiments}
This section will discuss the implementation of the four approaches that we presented earlier and the details of our experiment. We have shared our codebase and Colab notebooks on the public GitHub repository\cite{git}. Along with that, table \ref{table:1.2} lists all the names of the base models provided in the MLMAC\cite{mlmac} contest. Hence, all the experiments can be easily re-executed to reproduce the mentioned outcomes.
\begin{table}
\center
\begin{tabular}{ ||c|p{5cm}||  } 
 \hline
 \textbf{No. } & \multicolumn{1}{|c||}{\textbf{Model}} \\  
 \hline\hline

 1 & bloom-2b5\cite{bloom2b5}\\
 \hline
 2 & bloom-350m\cite{bloom350}\\
 \hline
 3 & Multilingual-MiniLM-L12-H384 \cite{wang2020minilm}\\
 \hline
 4 & distilgpt2\cite{sanh2019distilbert}\\
 \hline
 5 & codegen-350M-multi\cite{Nijkamp2022ACP}\\ 
 \hline
 6 &  opt-350m\cite{zhang2022opt}\\ 
 \hline
 7 & A gpt2-xl\cite{radford2019language}\\
 \hline
 8 &  gpt-neo-125M\cite{gpt-neo}\\
 \hline
 9 & xlnet-base-cased \cite{DBLP:journals/corr/abs-1906-08237}\\
 \hline
 10 & DialoGPT-large \cite{zhang2019dialogpt}\\ 
 \hline
 11 & gpt2 \cite{gpt2}\\
 \hline
 12 & gpt-j-6B \cite{mesh-transformer-jax}\\
 \hline
\end{tabular}
\caption{The table lists out names of all the base models provided in the MLMAC\cite{mlmac} contest.}
\label{table:1.2}
\end{table}

\subsection{EXPERIMENT-1: Machine Translation Evaluation metrics}
The machine translation evaluation metrics like BLEU\cite{papineni-etal-2002-bleu} and TER\cite{TER} scores are non-parametric scoring functions to measure the similarity in the provided reference and hypothesis text. Since these metrics do not have any trainable parameters, this approach does not require any pre-training for producing any inference. Therefore, we have used this approach directly for attributing the fine-tuned models with their corresponding base model. To begin the experiment, we first iterated over the provided list of twelve fine-tuned models and fed them ninety queries as input. We then fed the same queries to the list of twelve base models and recorded their responses. We have then iteratively compared the responses generated by each of the fine-tuned models and the base models via machine translation evaluation metric BLEU\cite{papineni-etal-2002-bleu} and TER\cite{TER} scores, where the responses generated by the fine-tuned models used as reference sentences and the responses produced by the base models utilized as hypotheses sentences. In the end, we have paired each of the fine-tuned models with the base model whose generated responses are most similar to its responses on average, as explained in algorithm \ref{alg1}. Since, at the time of the experiment, the correct base versions of the given fine-tuned models were unknown. Hence we have curated new dummy fine-tuned models for each of the provided base models to test the authenticity of the designed approach. Lastly, we repeated the entire simulation ten times and concluded an average accuracy of 7.1\%.

\subsection{EXPERIMENT-2: Vector Space Model}
In this experiment, we initially fed ninety queries to the provided twelve base models and recorded their responses. Then we constructed twelve documents for each of the twelve base models by accumulating their generated responses to develop the VSM\cite{vsm} model. After that, we fed the same queries to the fine-tuned models and used their responses to find the most suitable document for it with the help of cosine similarity. In the end, we have paired each fine-tuned model with the base model whose associated document was most suitable for its generated responses as postulated in algorithm \ref{alg2}. During the execution of this experiment, the organizers had not released the correct base versions for the provided fine-tuned models. Thus, to simulate the workability of this approach, we have developed new dummy fine-tuned models of our own. Since we know which dummy fine-tuned model originated from which of the base model, we can now identify the cases when the designed approach infers a wrong decision. Keeping all these mentioned details intact, we repeated the entire simulation ten times and recorded an average accuracy of 8.1\%.

\subsection{EXPERIMENT-3: Multiclass Text Classification}
We have curated two tabular datasets of 1080 records with three columns. The first tabular dataset is for training the multi-class text classifier, the second is for pairing the provided fine-tuned models with the given base models by using its inference. The first column of the training dataset contains the list of queries while the second and third column contains responses from the twelve base models against those queries, along with their names. There are a total of ninety distinct queries in our dataset, and the organizers of the MLMAC\cite{mlmac} competition have provided twelve base models. Therefore, recording a response from all the twelve base models against each of the ninety queries will yield a tabular dataset of (12*90) 1080 records. Likewise, the second tabular dataset contains similar queries in its first column while the second and third columns consist of responses from the twelve fine-tuned models against those queries, along with their names. The multi-class text classifier can use the training dataset to process the responses generated from the base models as its input and learn to identify which base model has emitted which response. After the training process completes, the developed multi-class text classifier can consume the responses of the fine-tuned models from the second tabular dataset and outputs the name for the most suitable base model that can generate the consumed response, as described in algorithm \ref{alg3}. During the execution of this experiment, the MLMAC\cite{mlmac} organizers haven't released the ground truth pairing of the fine-tuned models with their associated base versions. Hence, to test the effectiveness of this approach, we have applied stratified 10-fold cross-validation on the constructed training dataset. We have selected four different transformer-based multi-class text classifiers to execute this experiment. The table below postulates the name of these classifiers with the average accuracies they have achieved across the ten folds of the training dataset.
\begin{table}[h!]
\center
\begin{tabular}{ ||c|p{5cm}| p{1cm}|  } 
 \hline
 \textbf{No. } & \multicolumn{1}{|c|}{\textbf{Multiclass Text Classifier}} & \multicolumn{1}{|c||}{\textbf{Accuracy}}\\  
 \hline\hline

 1 & distilbert-base-uncased \cite{Sanh2019DistilBERTAD}& 9.3\%\\
 \hline
 2 & google/electra-small-discriminator \cite{clark2020electra}& 9.8\%\\
 \hline
 3 & google/albert-base-v2 \cite{DBLP:journals/corr/abs-1909-11942}& 9.5\%\\
  \hline
 4 & google/roberta-base \cite{DBLP:journals/corr/abs-1907-11692}& 10.1\%\\
  \hline
\end{tabular}
\caption{The table lists the names of the multi-class text classifiers that we have used to conduct the experiment, along with the average accuracies, they have achieved across the ten-folds of the training dataset.}
\label{table:2}
\end{table}
\subsection{EXPERIMENT-4:One-VS-All Multiclass Text Classification}
We have used the same training and inferencing dataset from the previous experiment to train the binary text classifiers via the one-vs-all technique. There are a total of twelve base models provided by MLMAC\cite{mlmac} organizers, and as we are utilizing the one-vs-all strategy, we have trained twelve binary text classifiers; one for each of the base models. Here, the only addition we made is that we have made twelve copies of the training data set, one for each binary text classifier. Then, we divided each copy of the data set into positive and negative examples to make binary text classifiers learn the responses their associated base model can produce and the responses it can not generate. We have used pre-trained DistilBERT implementation to construct these twelve binary text classifiers. After the training completes for all twelve binary text classifiers, we used the same inference dataset from the previous experiment that contains the responses of all the fine-tuned models. One by one, we fed the responses of the fine-tuned models as input to all the binary text classifiers and recorded their confidence scores. We have iteratively paired all responses of the fine-tuned models with the base model whose text classifier has emitted the highest score. In the end, we have paired each fine-tuned model with the base model that got paired most frequently with its responses, as mentioned in algorithm \ref{alg4}. At the time of setting this experiment, the MLMAC\cite{mlmac} organizers haven't released the correct pairings of the fine-tuned models with their base models. Therefore, to verify the usefulness of this approach, we have applied the stratified 10-fold cross-validation on the training set and obtained an average accuracy of 8.8\%.

\section{Conclusion}
Overall traditional multi-class text classification using the latest transformer-based BERT model has shown the best performance in our experiments compared to the other approaches for attributing fine-tuned models with their corresponding base models. In our experiments, we performed ten-fold cross-validation on the assembled training data set that did not contain any miscellaneous records or responses generated from additional third-party models provided by authorities other than MLMAC\cite{mlmac}. We strongly think that the absence of such miscellaneous records that don't belong to any of the models provided in the contest is the main reason behind the better performance of traditional multi-class text classification over the one-vs-all strategy using binary text classifiers. The vector space models have shown a comparable performance and a notable improvement over the translation evaluation metrics. Nonetheless, we have selected traditional multi-class text classification for generating the final submission that includes the pairing between provided fine-tuned and base models. Our submission ranked fourth, and we have also received an award of 2000\$ as a travel grant for attending the SaTML 2023 conference\cite{satmlIEEESaTML} in the US North Carolina to present our work. The results produced from our approach stood very close to the top submission on the leaderboard. The first submission correctly paired seven fine-tuned models out of twelve with their associated base versions using 1,212 queries. On the contrary, in our submission, we successfully paired six fine-tuned models with their corresponding base versions using 1,084 queries. The judges of the MLMAC\cite{mlmac} contest have assigned a higher rank to the submission that correctly paired more fine-tuned models with their base versions using the least number of queries. In that sense, the results produced by our approach were pretty close to the top submission. In the future, we strongly believe that constructing a logical sequence among the queries can yield further improved results with better accuracy. Deductive logic plays a significant role in real interrogation and inferring the most probable outcomes. But in our work, we have only randomly extracted queries from popular open-sourced datasets and have not attempted to build a logical sequence among them. That leads us to think that designing a process of sensibly choosing queries from open-sourced content and organizing them in a logical order can bring transforming results in this domain.

\section*{Acknowledgment}
The school of computing National University of Computer and Emerging Sciences \href{https://www.nu.edu.pk/}{(FAST-NU)} has provided immense support to back this research. We would like to acknowledge the efforts of the MLMAC\cite{mlmac} team to organize this contest and the SaTML\cite{satmlIEEESaTML} team for providing us travel grant to present our work on-site at the conference in the USA. We think the contest has provided a platform where students can positively compete with each other to test their skills and apply their Deep Learning knowledge. We will be glad to participate again in such events and hope these kinds of events should get organized more frequently. Personally, I also wanted to thank my supervisor, \href{https://www.linkedin.com/in/muhammad-rafi-34339219/}{Dr. Muhammad Rafi}, for providing specialized expertise and insights that greatly aided the research. Moreover, we are also thankful to \href{https://www.linkedin.com/in/hyrumanderson/}{Hyrum Anderson}, \href{https://scholar.google.com/citations?user=I2-QZyEAAAAJ&hl=en}{Keith Manville}, and \href{https://www.linkedin.com/in/deepeshvc/}{Deepesh Chaudhari} for keeping us posted about the updates of the challenge and solving our queries timely to help us throughout the event.

\bibliographystyle{IEEEtran}
\bibliography{ref}

\end{document}